# Forecasting Spoken Language Development in Children with Cochlear Implants Using Preimplantation MRI

Yanlin Wang[1#], Di Yuan[1,2#], Shani Dettman[3], Dawn Choo[3], Emily Shimeng Xu[4], Denise Thomas[5], Maura E Ryan[6], Patrick C M Wong[1,7*], Nancy M Young[4,8,9*]

[1]Brain and Mind institute, The Chinese University of Hong Kong, Hong Kong SAR, China

[2]Department of Psychology, The Chinese University of Hong Kong, Hong Kong SAR, China

[3]Department of Audiology & Speech Pathology, The University of Melbourne, 550 Swanston St, Parkville, Victoria 3010 Australia

[4]Division of Otolaryngology Head & Neck Surgery, Ann & Robert H. Lurie Children's Hospital of Chicago, Chicago, Illinois, United States.

[5]Department of Audiology, Ann & Robert H. Lurie Children's Hospital of Chicago, Chicago, Illinois, United States.

[6]Department of Medical Imaging, Ann & Robert H. Lurie Children's Hospital of Chicago, Chicago, Illinois, United States.

[7]Department of Linguistics and Modern Languages, The Chinese University of Hong Kong, Hong Kong SAR, China

[8]Department of Otolaryngology Head & Neck Surgery, Feinberg School of Medicine, Northwestern University, Chicago, Illinois, United States.

[9]Knowles Hearing Center, Department of Communication Sciences and Disorders, Northwestern University, Evanston, Illinois, United States.

\# These authors contributed equally to this work.

**\*Corresponding Authors:**

Nancy M Young, MD

Division of Otolaryngology, Ann & Robert H. Lurie Children's Hospital of Chicago

Department of Otolaryngology Head & Neck Surgery, Feinberg School of Medicine, Northwestern University

Knowles Hearing Center, Department of Communication Sciences and Disorders, Northwestern University

Email: nyoung@luriechildrens.org

Patrick C M Wong, PhD

Brain and Mind institute, The Chinese University of Hong Kong

Email: p.wong@cuhk.edu.hk

Cochlear implants (CI) significantly improve spoken language in children with severe-to-profound sensorineural hearing loss (SNHL), yet outcomes remain more variable than in children with normal hearing. This variability cannot be reliably predicted for individual children using age at implantation or residual hearing. This study aims to compare the accuracy of traditional machine learning (ML) to deep transfer learning (DTL) algorithms to predict post-CI spoken language development of children with bilateral SNHL using a binary classification model of high versus low language improvers. A total of 278 implanted children enrolled from three centers. The accuracy, sensitivity and specificity of prediction models based upon brain neuroanatomic features using traditional ML and DTL learning. DTL prediction models using bilinear attention-based fusion strategy achieved: accuracy of 92.39% (95% CI, 90.70%-94.07%), sensitivity of 91.22% (95% CI, 89.98%-92.47%), specificity of 93.56% (95% CI, 90.91%-96.21%), and area under the curve (AUC) of 0.977 (95% CI, 0.969-0.986). DTL outperformed traditional ML models in all outcome measures. DTL was significantly improved by direct capture of discriminative and task-specific information that are advantages of representation learning enabled by this approach over ML. The results support the feasibility of a single DTL prediction model for language prediction of children served by CI programs worldwide. (*Due to the notification of arXiv "The Abstract field cannot be longer than 1,920 characters", the appeared Abstract is shortened. For the full Abstract, please download the Article.)

Introduction

Cochlear implants (CI) are an effective treatment for young children with severe to profound sensorineural hearing loss (SNHL) that enables development of spoken language. [1] However, language outcomes after CI are variable in comparison to children with normal hearing. [2] Despite the availability of various early intervention approaches, there is little consensus on the optimal type and dose of behavioral therapy to improve listening and spoken language. [3] Accurate prediction of spoken language development on the individual child level prior to CI would allow for a customized "predict to prescribe" approach to reduce outcome variability. [4] Accurate prediction of spoken language improvement, in particular for parents of children likely to achieve lower language improvement, has potential to improve pre-CI parental counselling and post-CI therapy planning. Prediction gives parents and therapists the opportunity to arrange for more intensive behavioral therapy. Moreover, by forecasting the language developmental trajectory after CI, it becomes possible to evaluate the efficacy of different therapy approaches. Neural prediction may lead to development of more effective therapies based on pre-CI brain structure and function.

Brain measures serve as better prognostic indicators, either alone or in combination with other measures, than traditional measures such as age at implant and pre-implantation residual hearing. [5,6] Early auditory experience significantly impacts the development of auditory and language networks which are crucial for subsequent growth. [7-10] Studies have successfully used machine learning (ML) techniques to forecast the auditory and spoken language skills of children with CI. For example, using the pre-surgical neuroanatomical features from MRI and a ML linear support vector machine (SVM) classifier, prediction accuracy of 84% was achieved as to whether a child would experience high vs low speech perception improvement six months after device activation, as assigned by the group median outcome score. [11] In comparison, non-neural features, including age at implantation and residual hearing only reached a chance level of accuracy in predicting speech perception improvement. The robustness and efficiency of brain measures in predicting post-CI improvements in children and adults have also been supported by studies using brain imaging techniques to activate brain regions with audio and visual stimuli. [12-14]

Despite the increasing number of studies utilizing ML to predict post-CI outcomes, modeling brain data on a multicenter dataset remains challenging due to the variations in MRI scan protocol and outcome measurements. [15] Another complicating factor is that applying dimensional reduction—an essential preprocessing step for many ML algorithms—to heterogeneous datasets can oversimplify the data, leading to overfitting, reduced interpretability, and ultimately diminished model effectiveness. [16,17] Deep learning has shown considerable advantages in representation learning, the ability to automatically learn from useful data sources, and scalability over ML in modelling brain imaging data. [16] Deep transfer learning (DTL) can leverage prior knowledge learned from pretraining on a large dataset to enhance downstream task performance. [18,19] Multi-modal integration can identify feature value patterns across modalities (e.g., clinical findings such as age at implant and residual hearing; brain neuroanatomy imaging findings), which is essential as it impacts how well a model learns and generalizes across multiple data sources. [20,21] However, the variability in feature distributions and outcome measures across centers and languages fundamentally challenges the capability of deep learning to identify how information is encoded and processed by the brain, referred to as discriminative brain representations, for predicting CI improvement on a multicenter dataset. [22] Consequently, rigorous evaluation of the robustness of deep learning approaches on the multi-center dataset is imperative before deploying deep learning-based preoperative neural prediction models.

To this end, this study develops and compares neural predictive models for forecasting long-term post-CI improvements in children with CI and evaluates their performance and robustness on the multicenter dataset. The working hypothesis of the study is that neural prediction modelling done with DTL would be more accurate, sensitive and specific than traditional ML when applied to a heterogenous data set from CI programs located in different continents serving diverse populations.

Methods

**Participants**

Children with congenital or early onset SNHL who received cochlear implants between 2009 and 2022 were enrolled from three international centers: Chicago, United States; Melbourne, Australia; and Hong Kong, China. All the children underwent T1-weighted structural whole-brain magnetic resonance imaging (MRI) as a part of their pre-CI evaluation using each medical center's standard clinical protocol. Language evaluation was obtained pre- and post-CI for up to three years. This study was approved at each centre by the Joint Chinese University of Hong Kong – New Territories East Cluster Clinical Research Ethics Committee, the Stanley Manne Children's Research Institute's Institutional Review Board, and The Royal Children's Hospital, Human Research Ethics Committee.

As a study aiming to predict improvements in as many children with CI as possible, we imposed relatively broad inclusion/exclusion criteria. Children had to be from homes where the dominant family language is Cantonese (Hong Kong), English (Melbourne, Chicago), or Spanish (Chicago). Children with additional conditions known to affect language development (e.g. Down, Fragile X, autism spectrum disorder) independent of SNHL were excluded as were children with gross brain malformations. A total of 278 children were included. The demographic information is shown in Table 1. In addition, correlation matrix analyses of demographic variables were performed for each center (see supplementary materials).

**Clinical Measures**

Children's auditory skill, speech perception, receptive and/or expressive language abilities were measured before and up to 36 months after implantation using different assessment tools across centers (see supplementary materials for details of assessment tools).

We refer to all these measurements as 'spoken language' being aware that audition and speech perception are precursors for spoken language development. [23,24] Positive correlations have been demonstrated between speech perception and spoken language scores on standardized tests for children with hearing loss. [25,26] While variances could be introduced by differences in the assessment methods and timing, it is feasible to compare the spoken language ability across the centers and over time because of the heterotypic stability inherent in spoken language development. [27-30] Specifically, a child's spoken language rank-order in the population remains consistent across age as long as those characteristics share the same underlying construct and theoretical value. Therefore, instead of using the raw scores directly for fine-grained prediction, we separated the spoken language improvement into binary classifications (high-improvement and low-improvement) using a median split approach for children within each center.

The improvement of spoken language development from pre- to post-CI was quantified by the change of assessed scores as a function of assessment time for each participant. To this end, a linear mixed-effect model was constructed for each center with spoken language scores as the dependent variable, subject ID as a random intercept, as well as assessment time as a random slope. The fixed effects portion of the model included only the intercept term, as the influence of time on spoken language scores was captured in the random slope. The model can be expressed mathematically as Scores ~ 1 + (assessment time | subject ID). The random slope in the model allowed us to estimate individual differences in the rate of speech and language change over time. Children with slope values larger than the group median were labeled as 'high-improvement,' while those with slope values smaller than the group median were labeled as 'low-improvement.

**MRI acquisition and preprocessing**

The T1-weighted MRI image was obtained from each child before CI. The scanning parameters were optimized to obtain a good signal-to-noise ratio (Supplementary Material). MRI images were processed using the Advanced Normalization Tools (ANTs) in Python.[31] To increase the image quality, the images were resampled to 1 mm× 1 mm× 1 mm voxel size and pre-processed following the basic preprocessing pipeline for T1-weighted brain MRI in ANTs. The deformation-based morphometry (DBM) method was used to examine the morphological differences over the entire brain with an age appropriate T1 image as the template.[32,33] Fifteen axial 2D slices were extracted from the central part of the 3D DBM brain scans.[34] The images were cropped and resized into a target resolution of 128×128 voxels and were normalized using ImageNet statistics (mean=[0.485, 0.456, 0.406], std=[0.229, 0.224, 0.225]) before being passed on for further analyses.[35] Each slice was assigned the same label as the corresponding subject and used as a data sample to train the model. In addition, we conducted sensitivity analysis to assess potential bias arising from slice selection, evaluating model performance across different slice counts or positions. The detailed results of this analysis are provided in the supplementary materials.

**Transfer Learning and Feature Extractions**

Pre-trained convolutional neural network (CNN) models used included AlexNet,[36] VGG19,[37] ResNet,[38] GoogleNet,[39] Inception,[40] MobileNet,[41] and DenseNet,[42] implemented in PyTorch version 1.9, for feature extraction. This standard transfer learning strategy involves using pre-trained CNN models on ImageNet as the backbone of the model to capture generalizable features, followed by fine-tuning the top layers to learn new specialized representations tailored to our output classifier.[36,43] During the fine-tuning phase, the weights and biases of the CNN models were frozen to prevent changes. Subsequently, an attention-based fusion network was added to incorporate clinical measures into neural feature representations from the hidden layer's activation function to achieve a higher performance of the model by using a bilinear attention mechanism. Specifically, clinical measures included age at CI, age at MRI, age at hearing aid fitting, gender, left/right pure tone average residual hearing, and preoperative language ability scores. We utilized a bilinear attention network to capture high-level interactive relations among multiple modalities and then extracted joint image–meta representation by a bilinear pooling layer. [44,45] Data augmentation with random rotation and flipping was executed to improve the model training efficiency. [46,47] The loss function was binary cross-entropy with logit loss. The optimizer was Adam with a learning rate of 1×10-4. A total of 100 epochs with a batch size of 64 images were set for training. The validation performance was used to determine when to stop the training. The CNN models were trained until there was no improvement in the validation loss for 10 consecutive epochs. The model's performance was validated using five-fold cross-validation on 80% of the data, with the remaining 20% used as a held-out test set for evaluation.

**Performance comparisons**

To examine whether neural features can predict long-term post-CI improvements, we first compared state-of-the-art CNN models in the multi-center dataset. To evaluate whether our model's performance was robust to variations in data distribution across medical centers and languages, we evaluated the model performance on single datasets or combined datasets. Moreover, to improve the performance of the neural predictive model, we integrated clinical features with neural features by using a bilinear attention mechanism. In addition, to compare the effectiveness of DTL and traditional ML models in capturing shared and robust brain representations, we evaluated both approaches on a prediction task using a multi-center dataset. Seven DTL models and eight ML models–Lasso regression (LR), Ridge regression (RR), SVM, Random Forest (RF), Decision Tree (DT), K-Nearest Neighbor (KNN), and

eXtreme Gradient Boosting (XGBoost)- are compared. To reduce the dimensionality of the whole-brain voxel-wise features, we applied four linear and non-linear dimensionality-reduction techniques[16,48]: principal component analysis (PCA), Gaussian random projection (GRP), recursive feature elimination (RFE), and univariate feature selection (UFS). The detailed information on ML models and dimensionality reduction methods is provided in the Supplementary Information.

Results

Implanted children showed improvements in spoken language abilities compared to the baseline measurement tested before implantation (Fig 1). Specifically, in Chicago, the mean spoken language abilities of English-learning children improved from 75 to 292, and those of Spanish-learning children from 45 to 203, over the period from pre-CI to 36 months post-CI, as tested by SRI-m. Similarly, in Hong Kong, Cantonese-learning children showed an increase in mean scores from 17 to 32, over the period from pre-CI to 24 months post-CI, as measured by LittlEARS Auditory Questionnaire (range: 0-35). The rate of improvement was greatest over the first 1.5 years after initial implantation. In Melbourne, the mean receptive language of English-learning children improved from 74 to 85 in the first two years after implantation but dropped to 70 in the third year post-CI, as tested by the Picture Peabody Vocabulary Test-4 (PPVT) and Preschool Language Scale 4 and 5 (PLS-4&5) (standard score, mean: 100, standard deviation: 15). Please see more details in Table 1 and supplementary materials. The different pattern of changes in spoken language development may result from the standard scores obtained in Melbourne, which take age-appropriate normal-hearing children as a control, suggesting that children were able to catch up with their normal-hearing peers but still lagged behind in their long-term spoken language development. Despite different standardized tests being used to capture the spoken language development across the centers, our predictive models were constructed to only predict the binary classifications of low or high improvement.

Table 2 lists the DTL and standard ML models' training and testing of accuracy, sensitivities, specificities, and area under the curve (AUC). The results showed that DTL models can substantially improve the model's prediction performance compared to ML models with the UFS dimensionality reduction method (Fig 2A). Among the various deep learning CNN models, the MobileNet model exhibits the best performance with an accuracy

of 86.79% (95% CI, 85.398%-87.60%) and AUC of 0.924 (95% CI, 0.918-0.929) on the test dataset. In contrast, Ridge with UFS exhibited the superior performance with accuracy of 62.14% (95% CI, 59.25%-65.03%) and AUC of 0.621 (95% CI, 0.593-0.650) as compared with the other three dimensionality reduction approaches (Table S1). This indicated DTL models can learn both general and domain-specific feature representations through the pretrained and finetuning procedure yielding higher performance than standard ML models trained on lower-dimensional projections of high-dimensional inputs.

Moreover, regardless of whether a single dataset or a combination of different datasets was used to build the model, the MobileNet model demonstrated consistently accurate performance (Fig 2C). Specifically, it achieved an accuracy of 90.36% (95% CI, 89.96%-90.76%) and AUC of 0.947 (95% CI, 0.944-0.950) on the Chicago-English and Melbourne-English datasets (same language, two centers) and an accuracy of 89.48% (95% CI, 88.57%-90.39%) and AUC of 0.937 (95% CI, 0.934-0.939) on the Chicago-English and Chicago-Spanish datasets (same center, two languages). When tested on the Chicago, Melbourne, and Hong Kong datasets, it achieved an accuracy of 86.79% (95% CI, 85.98%-87.60%) and AUC of 0.924 (95% CI, 0.918-0.929).

To evaluate DTL and ML model performance on the type of data or format of data, referred to as modality, both single modality features and combined modality features were evaluated. Compared with the single modality model, the bilinear-attention based fused model achieved the best performance against other single models while the single neural model with MobileNet outperformed the single clinical model with logistic regression (Fig 2D and Table 3). Specifically, the bilinear attention-based fusion model demonstrates superior predictive performance, achieving an accuracy of 92.39% (95% CI, 90.70%-94.07%), and a high AUC of 0.977 (95% CI: 0.969-0.986). This significantly outperformed both baseline models: The model using only clinical measures showed limited predictive

utility, with an accuracy of 53.57% (95% CI, 50.86%-56.29%) and an AUC of 0.522 (95% CI: 0.489-0.555), indicating performance near chance level. The model using only neural features performed better than the clinical model with an accuracy of 86.79% (95% CI, 85.98%-87.60%) and an AUC of 0.924 (95% CI: 0.918-0.929), but remained substantially lower than the fusion model.

Discussion

In this multicenter study, we employed DTL on the preoperative neuroanatomical features obtained from presurgical MRI brain scans to predict up to 3-year spoken language improvements in children with CIs. Our models consistently demonstrated accurate performance in distinguishing between higher and lower improvement groups for both single dataset and combined datasets. A bilinear attention-based fusion model outperformed unimodal approaches by efficiently capturing cross-modal interactions between clinical characteristics and neural imaging features. Critically, our DTL approach demonstrated superior robustness and flexibility in predicting post-CI improvement from pre-CI neural data, effectively capturing discriminative and task-specific brain representations across multi-center and language datasets that the current ML methods are not able to match.

To our knowledge, this study represents the largest sample size ever used with brain measure to build a CI predictive model. [49-51] Our evaluation demonstrates that DTL achieves consistently higher accuracies through combined models, confirming robustness and flexibility to heterogeneous data. These experiments illustrated that DTL can extract the robust, shared feature representations obtained by each medical center from diverse populations. These findings suggest that inherent heterogeneity arising from factors such as scanner protocols and language outcomes necessitates explicit consideration during model training in multi-center studies to avoid characteristic-specific poor generalization. Furthermore, our results support the concept that preoperative neural features can predict post-CI improvements in children with diverse backgrounds, regardless of the specific assessment tools used.

The DTL approach has shown to be powerful in healthcare decisions for rare diseases, such as Alzheimer's disease,[52] cardiomyopathy,[53] diabetic retinopathy,[54] etc. Compared to a previous study by Geng et al that used voxel-based ML models to predict speech perception improvements six months post-CI with 37 children, [11] our study employing a DTL approach and using a larger sample size revealed a higher prediction accuracy even for long-term post-CI improvements. Critically, conventional voxel-based ML approaches appear limited in handling heterogeneous multi-center and language data to accurately predict long-term post-CI improvements. Moreover, dimensionality reduction techniques used to reduce the number of features in a dataset while preserving essential information often fail to extract shared representations from multi-center datasets, fundamentally limiting their effectiveness compared with DTL methods that can exploit such variability.

Deep learning methods demonstrate significant advantages over traditional ML approaches in harnessing large, heterogeneous datasets, especially when paired with transfer learning. [55] For example, Abrol et al. reported that deep learning consistently improved performance at larger training datasets on neuroimaging classification and regression tasks as sample size increased. [56] While deep learning approaches benefit from larger sample sizes, the inherent heterogeneity of multi-center data necessitates careful handling of representation learning and model development to improve model performance. In our case, transfer learning enabled our models to capture robust, discriminative brain representations, achieving an 87% accuracy for post-CI outcome prediction on the combined dataset. Furthermore, our novel bilinear attention-based fusion network effectively integrated clinical measures with neural features, significantly enhancing preoperative prediction accuracy to 92.39%. These findings demonstrate that, beyond sample size, effectively leveraging inherent data heterogeneity and multiple modalities is critical for improving model performance and robustness in preoperative neural prediction tasks.

Our study had several limitations. First, the need to accommodate the different outcome measures across centers by use of binary classifications (high improvement and low improvement) using a median split method limits differentiation of children with medium-level outcome measures. Second, processing each 2D slice independently reduces the spatial information between slices. This was mitigated by using transfer learning and fine-tuning techniques to integrate prior knowledge from large datasets with domain-specific knowledge. Nevertheless, slice-based approach remains suboptimal for modeling complex volumetric patterns, underscoring the need for future work to incorporate explainable AI techniques. Third, the high performance observed in the present study may partly reflect the cohort-specific factors and the enlarged number of image inputs inherent to the slice-based approach.[57] To mitigate this potential overfitting, we employed cross-validation and multiple regularization strategies including dropout, weight decay, and early stopping. Finally, robustness is a prerequisite for deep learning algorithms to generalize across centers.[17] Although our experiments have demonstrated the robustness of DTL in modelling brain data, cross-center generalization was limited due to variations in features and outcomes across centers. Future research should focus on testing the model's generalizability across diverse populations and implant programs worldwide.

Conclusions

Our study demonstrated the robustness of the DTL approach for neural prediction of whether children will have high or low spoken language improvement after CI. Furthermore, our model provides more accurate preoperative prediction by employing techniques that leverage data from multiple sources to improve performance. This study supports the feasibility of the development of a single accurate DTL neural prediction model to use across centers and languages worldwide. Accurate prediction of spoken language on the individual child level is an important first step in the creation of customized treatment plans to optimize language after implantation.

**Acknowledgements**

This work was supported by the Research Grants Council of Hong Kong Grant GRF14605119, National Institutes of Health R21DC016069 and R01DC019387. The authors thank Michael C. F. Tong, Iris H.-Y. Ng, Wai Tsz Chang, Claire Iseli, Denise Courtenay, Mariz Hanna, and Robert Briggs.

**Conflict of Interest Disclosures**

Nancy M. Young and Patrick C. M. Wong declare that they are the inventors of an invention that has been granted a patent (US11607309B2) that is related to the research reported here. Nancy M. Young receives research funding from MEDEL for an FDA clinical trial and participated in a US MEDEL surgical advisory meeting in 2025. No other disclosures were reported.

**Availability of data and code**

The datasets in the current study are not publicly available due to strict privacy regulations set forth by the Institutional Review Board. All code of machine learning and deep transfer learning models can be found at https://github.com/DLDLCQJ/Cochlear-Implant-Identification.

# References

1. Sharma SD, Cushing SL, Papsin BC, Gordon KA. Hearing and speech benefits of cochlear implantation in children: A review of the literature. *International journal of pediatric otorhinolaryngology*. 2020;133:109984.
2. Niparko JK, Tobey EA, Thal DJ, et al. Spoken language development in children following cochlear implantation. *Jama*. 2010;303(15):1498-1506.
3. Chu C, Dettman S, Choo D. Early intervention intensity and language outcomes for children using cochlear implants. *Deafness & Education International*. 2020;22(2):156-174.
4. Wong PCM, Vuong LC, Liu K. Personalized learning: From neurogenetics of behaviors to designing optimal language training. *Neuropsychologia*. Apr 2017;98:192-200. doi:10.1016/j.neuropsychologia.2016.10.002
5. Gabrieli JD, Ghosh SS, Whitfield-Gabrieli S. Prediction as a humanitarian and pragmatic contribution from human cognitive neuroscience. *Neuron*. 2015;85(1):11-26.
6. Yuan D, Chang WT, Ng IH-Y, et al. Predicting Auditory Skill Outcomes After Pediatric Cochlear Implantation Using Preoperative Brain Imaging. *American Journal of Audiology*. 2025;34(1):51-59.
7. Perani D, Saccuman MC, Scifo P, et al. Neural language networks at birth. *Proceedings of the National Academy of Sciences*. 2011;108(38):16056-16061. doi:doi:10.1073/pnas.1102991108
8. Dehaene-Lambertz G, Hertz-Pannier L, Dubois J, et al. Functional organization of perisylvian activation during presentation of sentences in preverbal infants. *Proc Natl Acad Sci U S A*. Sep 19 2006;103(38):14240-5. doi:10.1073/pnas.0606302103
9. Yuan D, Ng IH, Feng G, et al. The Extent of Hearing Input Affects the Plasticity of the Auditory Cortex in Children With Hearing Loss: A Preliminary Study. *Am J Audiol*. Jun 2023;32(2):379-390. doi:10.1044/2023_aja-22-00172
10. Yuan D, Tournis E, Ryan ME, et al. Early-stage use of hearing aids preserves auditory cortical structure in children with sensorineural hearing loss. *Cereb Cortex*. Apr 1 2024;34(4)doi:10.1093/cercor/bhae145
11. Feng G, Ingvalson EM, Grieco-Calub TM, et al. Neural preservation underlies speech improvement from auditory deprivation in young cochlear implant recipients. *Proceedings of the National Academy of Sciences*. 2018;115(5):E1022-E1031.
12. Lu S, Xie J, Wei X, et al. Machine learning-based prediction of the outcomes of cochlear implantation in patients with cochlear nerve deficiency and normal cochlea: a 2-year follow-up of 70 children. *Frontiers in Neuroscience*. 2022;16:895560.
13. Song Q, Qi S, Jin C, et al. Functional brain connections identify sensorineural hearing loss and predict the outcome of cochlear implantation. *Frontiers in Computational Neuroscience*. 2022;16:825160.
14. Tan L, Holland SK, Deshpande AK, Chen Y, Choo DI, Lu LJ. A semi-supervised support vector machine model for predicting the language outcomes following cochlear implantation based on pre-implant brain fMRI imaging. *Brain and Behavior*. 2015;5(12):e00391.
15. Kim H, Kang WS, Park HJ, et al. Cochlear implantation in postlingually deaf adults is time-sensitive towards positive outcome: prediction using advanced machine learning techniques. *Scientific reports*. 2018;8(1):18004.
16. Abrol A, Fu Z, Salman M, et al. Deep learning encodes robust discriminative neuroimaging representations to outperform standard machine learning. *Nature communications*. 2021;12(1):353.
17. Goetz L, Seedat N, Vandersluis R, van der Schaar M. Generalization—a key challenge for responsible AI in patient-facing clinical applications. *npj Digital Medicine*. 2024;7(1):126.
18. Huh M, Agrawal P, Efros AA. What makes ImageNet good for transfer learning? *arXiv preprint arXiv:160808614*. 2016;
19. Zhao Z, Alzubaidi L, Zhang J, Duan Y, Gu Y. A comparison review of transfer learning and self-supervised learning: Definitions, applications, advantages and limitations. *Expert Systems with Applications*. 2024;242:122807.

20. Wang Y, Tang S, Ma R, Zamit I, Wei Y, Pan Y. Multi-modal intermediate integrative methods in neuropsychiatric disorders: A review. *Comput Struct Biotechnol J*. 2022;20:6149-6162. doi:10.1016/j.csbj.2022.11.008
21. Picard M, Scott-Boyer M-P, Bodein A, Périn O, Droit A. Integration strategies of multi-omics data for machine learning analysis. *Computational and Structural Biotechnology Journal*. 2021;19:3735-3746.
22. Balendran A, Beji C, Bouvier F, et al. A scoping review of robustness concepts for machine learning in healthcare. *npj Digital Medicine*. 2025;8(1):38.
23. Perigoe CB, Paterson MM. Understanding auditory development and the child with hearing loss. *Fundamentals of audiology for the speech-language pathologist*. 2013:173-204.
24. Werker JF, Hensch TK. Critical periods in speech perception: new directions. *Annual review of psychology*. 2015;66(1):173-196.
25. Geers AE, Nicholas JG, Sedey AL. Language skills of children with early cochlear implantation. *Ear and hearing*. 2003;24(1):46S-58S.
26. DesJardin JL, Ambrose SE, Martinez AS, Eisenberg LS. Relationships between speech perception abilities and spoken language skills in young children with hearing loss. *International journal of audiology*. 2009;48(5):248-259.
27. Bornstein MH, Putnick DL, Esposito G. Continuity and stability in development. *Child development perspectives*. 2017;11(2):113-119.
28. Bornstein MH, Hahn C-S, Putnick DL, Pearson RM. Stability of core language skill from infancy to adolescence in typical and atypical development. *Science Advances*. 2018;4(11):eaat7422.
29. Wong PCM, Lai CM, Chan PHY, et al. Neural Speech Encoding in Infancy Predicts Future Language and Communication Difficulties. *Am J Speech Lang Pathol*. Sep 23 2021;30(5):2241-2250. doi:10.1044/2021_ajslp-21-00077
30. Novitskiy N, Maggu AR, Lai CM, et al. Early Development of Neural Speech Encoding Depends on Age but Not Native Language Status: Evidence From Lexical Tone. *Neurobiol Lang (Camb)*. 2022;3(1):67-86. doi:10.1162/nol_a_00049
31. Tustison NJ, Cook PA, Klein A, et al. Large-scale evaluation of ANTs and FreeSurfer cortical thickness measurements. *Neuroimage*. Oct 1 2014;99:166-79. doi:10.1016/j.neuroimage.2014.05.044
32. Gaser C, Nenadic I, Buchsbaum BR, Hazlett EA, Buchsbaum MS. Deformation-based morphometry and its relation to conventional volumetry of brain lateral ventricles in MRI. *NeuroImage*. 2001;13(6):1140-1145.
33. Shi F, Yap P-T, Wu G, et al. Infant brain atlases from neonates to 1-and 2-year-olds. *PloS one*. 2011;6(4):e18746.
34. Wen J, Thibeau-Sutre E, Diaz-Melo M, et al. Convolutional neural networks for classification of Alzheimer's disease: Overview and reproducible evaluation. *Medical image analysis*. 2020;63:101694.
35. Ardalan Z, Subbian V. Transfer learning approaches for neuroimaging analysis: a scoping review. *Frontiers in Artificial Intelligence*. 2022;5:780405.
36. Krizhevsky A, Sutskever I, Hinton GE. Imagenet classification with deep convolutional neural networks. *Advances in neural information processing systems*. 2012;25
37. Simonyan K. Very deep convolutional networks for large-scale image recognition. *arXiv preprint arXiv:14091556*. 2014;
38. He K, Zhang X, Ren S, Sun J. Deep residual learning for image recognition. 2016:770-778.
39. Szegedy C, Liu W, Jia Y, et al. Going deeper with convolutions. 1-9.
40. Szegedy C, Vanhoucke V, Ioffe S, Shlens J, Wojna Z. Rethinking the Inception Architecture for Computer Vision. *2016 IEEE Conference on Computer Vision and Pattern Recognition (CVPR)*. 2015:2818-2826.
41. Howard AG. Mobilenets: Efficient convolutional neural networks for mobile vision applications. *arXiv preprint arXiv:170404861*. 2017;
42. Huang G, Liu Z, Van Der Maaten L, Weinberger KQ. Densely connected convolutional networks. 2017:4700-4708.
43. Yosinski J, Clune J, Bengio Y, Lipson H. How transferable are features in deep neural networks? *Advances in neural information processing systems*. 2014;27

44. Kim J-H, Jun J, Zhang B-T. Bilinear attention networks. *Advances in neural information processing systems*. 2018;31
45. Bai P, Miljković F, John B, Lu H. Interpretable bilinear attention network with domain adaptation improves drug–target prediction. *Nature Machine Intelligence*. 2023;5(2):126-136.
46. Taqi AM, Awad A, Al-Azzo F, Milanova M. The impact of multi-optimizers and data augmentation on TensorFlow convolutional neural network performance. IEEE; 2018:140-145.
47. Afzal S, Maqsood M, Nazir F, et al. A data augmentation-based framework to handle class imbalance problem for Alzheimer's stage detection. *IEEE access*. 2019;7:115528-115539.
48. Salam MA, Azar AT, Elgendy MS, Fouad KM. The effect of different dimensionality reduction techniques on machine learning overfitting problem. *Int J Adv Comput Sci Appl*. 2021;12(4):641-655.
49. Crowson MG, Lin V, Chen JM, Chan TCY. Machine Learning and Cochlear Implantation-A Structured Review of Opportunities and Challenges. *Otol Neurotol*. Jan 2020;41(1):e36-e45. doi:10.1097/mao.0000000000002440
50. Mo JT, Chong DS, Sun C, Mohapatra N, Jiam NT. Machine-Learning Predictions of Cochlear Implant Functional Outcomes: A Systematic Review. *Ear Hear*. Jul-Aug 01 2025;46(4):952-962. doi:10.1097/aud.0000000000001638
51. You E, Lin V, Mijovic T, Eskander A, Crowson MG. Artificial Intelligence Applications in Otology: A State of the Art Review. *Otolaryngol Head Neck Surg*. Dec 2020;163(6):1123-1133. doi:10.1177/0194599820931804
52. Saleh AW, Gupta G, Khan SB, Alkhaldi NA, Verma A. An Alzheimer's disease classification model using transfer learning Densenet with embedded healthcare decision support system. *Decision Analytics Journal*. 2023;9:100348.
53. Theodoris CV, Xiao L, Chopra A, et al. Transfer learning enables predictions in network biology. Nature. 2023;618(7965):616-624.
54. Dai L, Wu L, Li H, et al. A deep learning system for detecting diabetic retinopathy across the disease spectrum. Nature communications. 2021;12(1):3242.
55. Mahmood U, Rahman MM, Fedorov A, et al. Whole MILC: generalizing learned dynamics across tasks, datasets, and populations. Springer; 2020:407-417.
56. Shafieibavani E, Goudey B, Kiral I, et al. Predictive models for cochlear implant outcomes: Performance, generalizability, and the impact of cohort size. Trends in Hearing. 2021;25:23312165211066174.
57. Zhao Z, Chuah JH, Lai KW, et al. Conventional machine learning and deep learning in Alzheimer's disease diagnosis using neuroimaging: A review. *Front Comput Neurosci*. 2023;17:1038636. doi:10.3389/fncom.2023.1038636

**Table 1.** Demographic information for participants from different centers.

| | Chicago data | | Melbourne data | Hong Kong data | All |
|---|---|---|---|---|---|
| Sample size | 143 | 37 | 81 | 17 | 278 |
| Family language | English | Spanish | English | Cantonese | NA |
| Female, No. (%) | 67 (46.9) | 21 (56.8) | 37 (45.7) | 12 (70.6） | 137 (49.3) |
| Age at SNHL diagnosis, mean (SD), mo | 10.2 (13.3) | 11.1 (12.4) | 3.2 (4.4) | 11.6 (15.2) | 9.7 (12.8) |
| Age of HA fitting, mean (SD), mo | 11.6 (13.2) | 12.3 (12.5) | 3.8 (4.2) | 16.9 (13.6) | 10.4 (12.3) |
| Age at MRI, mean (SD), mo | 23.8 (20.5) | 26.9 (18.2) | 11.4 (12.1) | 24.3 (18.0) | 20.7 (18.9) |
| Age at CI, mean (SD), mo | 27.4 (20.9) | 30.1 (18.4) | 19.2 (13.2) | 32.5 (16.6) | 25.7 (18.8) |
| Unaided hearing of left ear, mean (SD), dB HL | 95.4 (17.0) | 98.9 (18.0) | 97.7 (18.7) | 103.3 (15.7) | 96.9 (17.5) |
| Unaided hearing of right ear, mean (SD), dB HL | 93.7 (18.1) | 100.2 (15.1) | 99.5 (19.0) | 101.7 (14.0) | 96.5 (17.9) |
| SES, mean (SD), | 86050.17 (31011.63) | 63300.58 (14334.92) | 1352 (679.98) | unknown | NA |
| Device Manufacturer | Cochlear Americas (71) | Cochlear Americas (24) | Cochlear Americas | Cochlear Americas | NA |
| | Advanced Bionics (24) | Advanced Bionics (6) | | | |
| | Med-EI (48) | Med-EI (7) | | | |
| Device Configuration | CI-Sequential (53) | CI-Sequential (21) | CI-Sequential (22) | CI-Simultaneous (7) | NA |
| | CI-Simultaneous (51) | CI-Simultaneous (3) | CI-Simultaneous (36) | CI-Unilateral (1) | |
| | CI-Unilateral (10) | CI-Unilateral (8) | unknown (23) | Bi-Model (1) | |
| | Bi-Model (29) | Bi-Model (5) | | unknown (8) | |
| *entire group* | | | | | |
| pre-CI (mean,SD) | 75.09 (114.23) | 45.58 (89.58) | 74.21 (16.28) | 16.81 (10.44) | NA |
| 6 mon post-CI (mean, SD) | 145.07 (118.51) | 93.76 (90.09) | Not tested | 22.83 (8.91) | NA |
| 12 mon post-CI (mean,SD) | 177.34 (134.16) | 136.41 (106.98) | 81.30 (20.29) | 29.00 (4.20) | NA |
| 18 mon post-CI (mean, SD) | 223.19 (139.10) | 192.76 (126.44) | Not tested | / | NA |
| 24 mon post-CI (mean, SD) | 249.17 (132.94) | 201.70 (134.79) | 84.68 (21.37) | 32.00 (2.73) | NA |
| 36 mon post-CI (mean, SD) | 291.50 (128.87) | 203.13 (110.49) | 69.6 (16.29) | / | NA |
| *Low improvement group* | | | | | |
| pre-CI (mean,SD) | 71.76 (119.41) | 47.66 (101.77) | 72.12 (15.77) | 24.75 (7.25) | NA |
| 6 mon post-CI (mean, SD) | 119.75 (108.41) | 84.68 (98.43) | Not tested | 28.50 (5.61) | NA |
| 12 mon post-CI (mean,SD) | 133.48 (114.73) | 110.46 (109.28) | 68.66 (13.85) | 32.6 (2.70) | NA |
| 18 mon post-CI (mean, SD) | 136.24 (99.89) | 132.31 (115.13) | Not tested | Not tested | NA |
| 24 mon post-CI (mean, SD) | 169.85 (134.27) | 137.60 (120.02) | 76.68 (10.93) | 31.25 (3.30) | NA |
| 36 mon post-CI (mean, SD) | 125 (NA) | 125 (NA[†]) | 62 (12.22) | Not tested | NA |
| *High improvement group* | | | | | |
| pre-CI (mean,SD) | 78.51 (109.40) | 43.26 (76.76) | 76.14 (16.72) | 8.88 (6.10) | NA |
| 6 mon post-CI (mean, SD) | 170.38 (123.65) | 103.98 (81.64) | Not tested | 17.17 (8.13) | NA |
| 12 mon post-CI (mean,SD) | 212.57 (139.04) | 164.52 (101.41) | 93.95 (17.73) | 26.43 (2.99) | NA |
| 18 mon post-CI (mean, SD) | 276.94 (133.15) | 268.31 (100.23) | Not tested | Not tested | NA |
| 24 mon post-CI (mean, SD) | 288.83 (114.08) | 272.92 (117.60) | 98.04 (17.75) | 32.75 (2.22) | NA |
| 36 mon post-CI (mean, SD) | 315.29 (118.72) | 281.25 (NA[†]) | 87.33 (8.50) | Not tested | NA |

Abbreviations: CI, cochlear implant; MRI, magnetic resonance imaging; HA, hearing aid; SNHL, sensorineural hearing loss; NA, not applicable or not available; SES, socioeconomic status; [†]only one data was available, the standard deviation cannot be calculated.

**Table 2.** Performance comparison of Deep Transfer Learning and Machine Learning models on multi-center dataset

| Types | Models | % (95% CI) | | | AUC (95% CI) |
|---|---|---|---|---|---|
| | | Accuracy | Sensitivity | Specificity | |
| Slice-based | VGG19-bn | 78.15 (76.41-79.90) | 76.00 (74.38-77.62) | 80.09 (77.84-82.34) | 0.866 (0.778-0.823) |
| | ResNet-50d | 76.68 (74.84-78.52) | 76.63 (72.71-80.55) | 76.60 (70.48-82.72) | 0.855 (0.844-0.867) |
| | DenseNet-169 | 86.14 (85.80-86.48) | 85.22 (84.43-86.00) | 86.97 (86.05-87.90) | 0.902 (0.900-0.905) |
| | AlexNet | 78.15 (77.41-78.89) | 82.55 (79.90-85.20) | 75.59 (73.36-77.82) | 0.841 (0.834-0.847) |
| | Inceptio-V3 | 75.24 (74.34-76.14) | 77.19 (71.96-82.42) | 73.12 (66.60-79.64) | 0.829 (0.821-0.838) |
| | GoogleNet | 81.10 (79.81-82.40) | 81.52 (79.65-83.39) | 80.73 (79.02-82.43) | 0.870 (0.867-0.872) |
| | MobileNet | 86.79 (85.98-87.60) | 89.90 (88.18-91.63) | 83.74 (81.24-86.25) | 0.924 (0.918-0.929) |
| Voxel-based[b] | Lasso | 58.57 (53.76-63.38) | 51.43 (44.01-58.85) | 65.71 (61.75-69.68) | 0.586 (0.538-0.634) |
| | Ridge | 62.14 (59.25-65.03) | 55.72 (47.07-64.36) | 68.57 (61.28-75.59) | 0.621 (0.593-0.650) |
| | DT | 60.71 (56.57-64.86) | 42.86 (31.99-53.72) | 78.57 (64.20-92.94) | 0.607 (0.566-0.649) |
| | SVM | 60.36 (57.47-63.25) | 55.71 (53.29-58.14) | 65.00 (59.22-70.78) | 0.604 (0.575-0.632) |
| | KNN | 59.64 (54.30-64.98) | 56.43 (45.84-67.02) | 62.86 (54.80-70.91) | 0.596 (0.543-0.650) |
| | RF | 59.64 (54.30-64.98) | 47.86 (38.66-57.05) | 71.43 (68.29-74.56) | 0.596 (0.543-0.650) |
| | Xgboost | 59.64 (49.65-69.63) | 55.00 (42.22-67.78) | 64.29 (51.36-77.21) | 0.596 (0.497-0.696) |

[a]Abbreviations: LR, Logistic Regression; KNN, K-Nearest Neighbor; SVM, Support Vector Machine; DT, Decision Tree; RT, Random Forest; XGBoost, eXtreme Gradient Boosting.

[b]UFS feature extraction

**Table 3.** Model performance comparison across datasets and modalities.

| Datasets | % (95% CI) | | | AUC(95% CI) |
|---|---|---|---|---|
| | **Accuracy** | **Sensitivity** | **Specificity** | |
| Subsets | | | | |
| Lure_Eng | 89.56 (89.18-89.93) | 92.02 (90.75-93.29) | 86.89 (85.69-88.10) | 0.943 (0.941-0.946) |
| Melb_Eng | 90.62 (89.86-91.38) | 91.83 (90.30-93.37) | 89.43 (88.72-90.14) | 0.949 (0.946-0.953) |
| Lure_Span | 90.81 (88.66-92.96) | 95.41 (93.71-97.11) | 85.20 (81.04-89.36) | 0.977 (0.968-0.986) |
| Lure_Eng_Span | 89.48 (88.57-90.39) | 88.64 (88.19-89.04) | 90.29 (88.90-91.67) | 0.937 (0.934-0.939) |
| Lure_Eng_Melb | 90.36 (89.96-90.76) | 90.65 (88.62-92.69) | 90.03 (87.89-92.17) | 0.947 (0.944-0.950) |
| Lure_Eng_Span_Melb | 86.00 (85.32-86.68) | 81.29 (79.60-82.97) | 90.29 (88.85-91.74) | 0.904 (0.895-0.914) |
| All Dataset[a] | | | | |
| Clinical_Features | 53.57 (50.86-56.29) | 52.41 (44.18-60.65) | 54.81 (47.26-62.37) | 0.522 (0.489-0.555) |
| Neural_Features | 86.79 (85.98-87.60) | 89.90 (88.18-91.63) | 83.74 (81.24-86.25) | 0.924 (0.918-0.929) |
| Clinical Features + Neural Features | 92.39 (90.70-94.07) | 91.22 (89.98-92.47) | 93.56 (90.91-96.21) | 0.977 (0.969-0.986) |

[a]All dataset comprise Chicago (English and Spanish), Melbourne, and Hong Kong.

**Figure Ledges:**

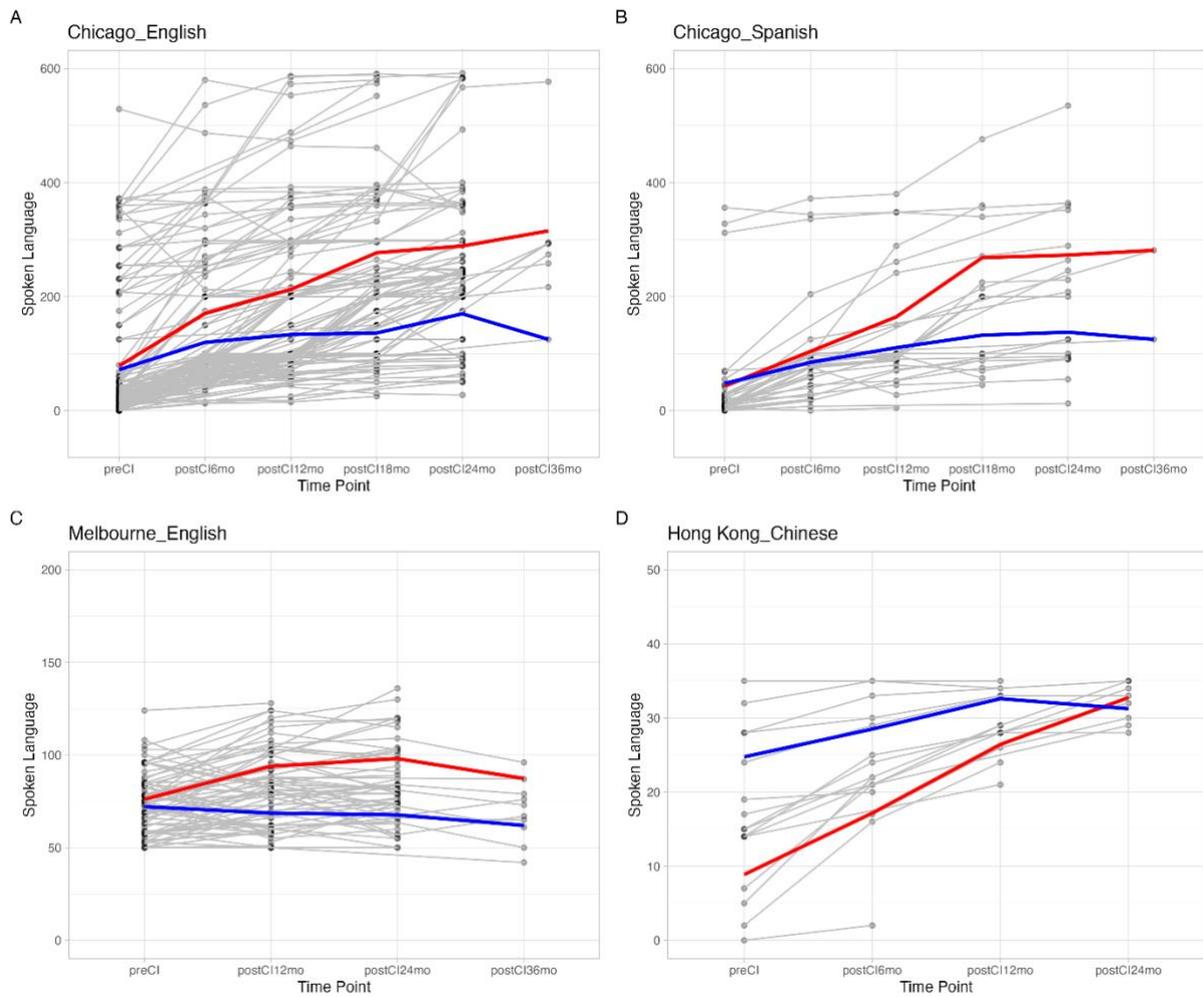

**Figure1**. Spoken language ability of children from before to after implantation at each center. A, Chicago English group. The spoken language ability was measured using Recognition Index-modified version (SRI-m), a hierarchical battery ranged from 0 to 600. B, Chicago Spanish group. The spoken language ability was measured using SRI-m Spanish version. C, Melbourne English group. The spoken language ability was assessed using two norm-referenced instruments—the Picture Peabody Vocabulary Test-4 (PPVT-4) and Preschool Language Scale 4 and 5 (PLS-4&5). Standard scores where 100 (+/- 15 is the normative mean). D, Hong Kong Chinese group. LittlEARS Auditory questionnaire was used with the score ranged from 0 to 35. Each grey line represents the spoken language development trajectory for one child. The red line represents the mean spoken language score for the high-improvement group. The blue line represents the mean spoken language score for the low-improvement group.

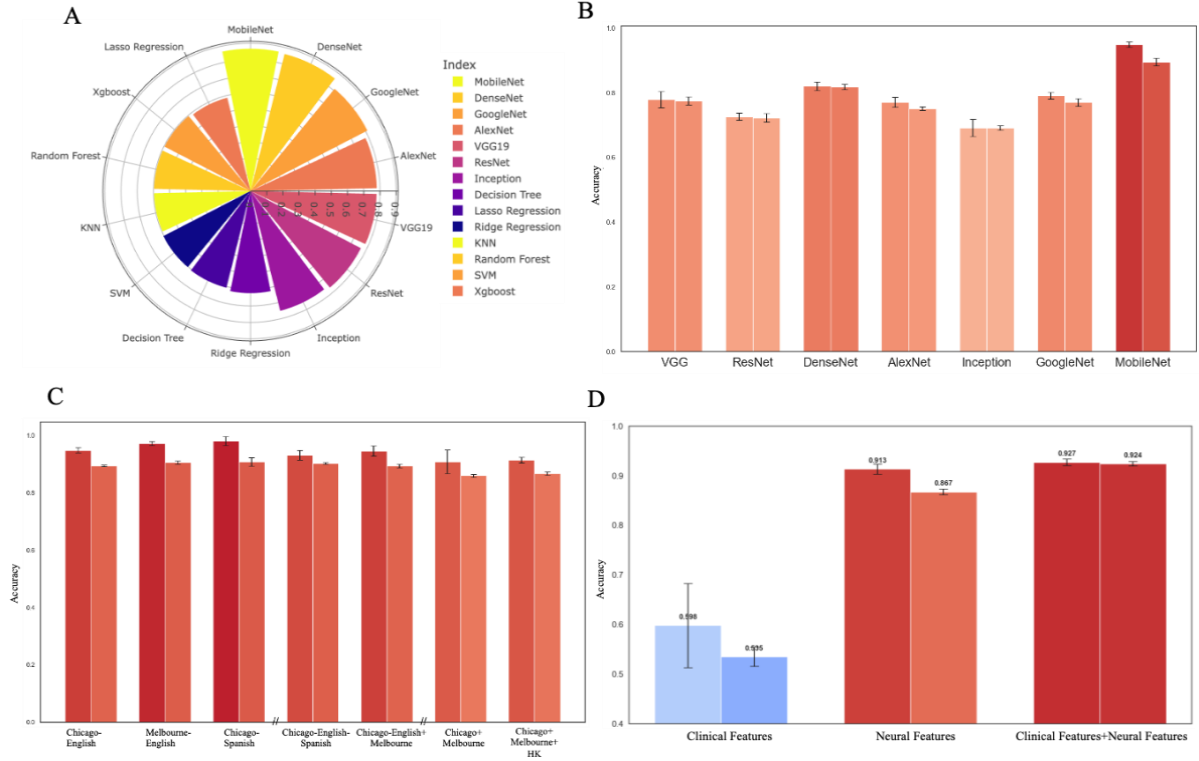

**Figure2**. Exhaustive evaluation of model performance comparison across datasets and modalities. A, Overall comparison of machine learning models versus deep transfer learning models on the multi-site dataset. B, Performance comparison among different transfer learning models on the multi-site dataset. C, Evaluation of deep transfer learning model separately on the Chicago-English dataset, across languages or centers cohort, and on the combined dataset. D, Performance comparison between the fused model and single models.

Supplementary Materials

**Spoken language measurements**

*Chicago data.* The spoken language ability was assessed using the Speech Recognition Index-modified version (SRI-m) before implantation and at 6, 12, 18, 24, and 36 months after CI. SRI-m is a hierarchical battery of age-appropriate measures, which assesses the speech recognition abilities from parental reports for children with lower auditory abilities to direct measures of speech perception for children with higher auditory abilities. The SRI-m has been used in a Childhood Development after Cochlear Implantation (CDaCI) Study.[1,2] It consists of the Infant-Toddler Meaningful Auditory Integration Scale/Meaningful Auditory Integration Scale (IT-MAIS/ MAIS), Early Speech Perception Test (ESP), Multisyllabic Lexical Neighborhood Test/Lexical Neighborhood Test (M/LNT), the Phonetically Balanced Word Lists-Kindergarten (PBK), the Pediatric Az Bio in quiet (AzBio-q), and the Pediatric Az Bio in noise (AzBio-n). The clinician decided which test the children should be tested with based on their age, developmental ability, and hearing aptitudes. To reflect children's spoken language development on the same scale, the scores of these tests were rescaled resulting in a final score ranging from 0 to 600 with higher scores representing better spoken language abilities.

*Melbourne data.* The ability of receptive and expressive language was assessed using two norm-referenced instruments—Pre-school Language Scale [PLS-4, PLS-5] [4] and Peabody Picture Vocabulary Test [PPVT-Revised, 3rd and 4th Editions] [3] before implantation and at 12, 24, and 36 months after CI. The tools were chosen to be appropriate for the age and stage for each child. The PLS-4 and PLS-5 assesses children from birth to 6 years 11 months and birth to 7 years 11 months, respectively, through tasks administered by the clinician using toys and picture-based materials. Standard scores and age equivalents are derived for receptive, expressive, and total language. The PPVT measures receptive vocabulary for children aged 2 years 6 months through to adulthood. The child hears a stimulus word and selects the picture that best represents it from four options. Standard scores and age equivalents are derived for receptive vocabulary. The two assessment tools offer age-based standard scores with a mean of 100 and a standard deviation of 15.

*Hong Kong data*. The spoken language ability of the children was evaluated using the LittlEARS Auditory Questionnaire, which relies on caregivers' observation of children's auditory behaviors (including that of spoken language) in daily life.[5] This 35-item questionnaire, with 'yes' or 'no' responses, takes about 10 minutes for caregivers to complete. The LittlEARS has been validated in children with both normal hearing and hearing loss.[6,7] Specifically, the questionnaire has demonstrated evidence of validity in accounting for the substantial variation in spoken language development among children with CIs.[8,9] In this study, the children's spoken language scores as measured by LittlEARS were obtained before CI and at 6, 12, and 24 months after CI.

**Statistical Analyses**

To improve the applicability of the model, we provided more information at each center that consistently contributes to outcomes. Correlation analyses were conducted illustrating relationships between behavioral factors at each center.

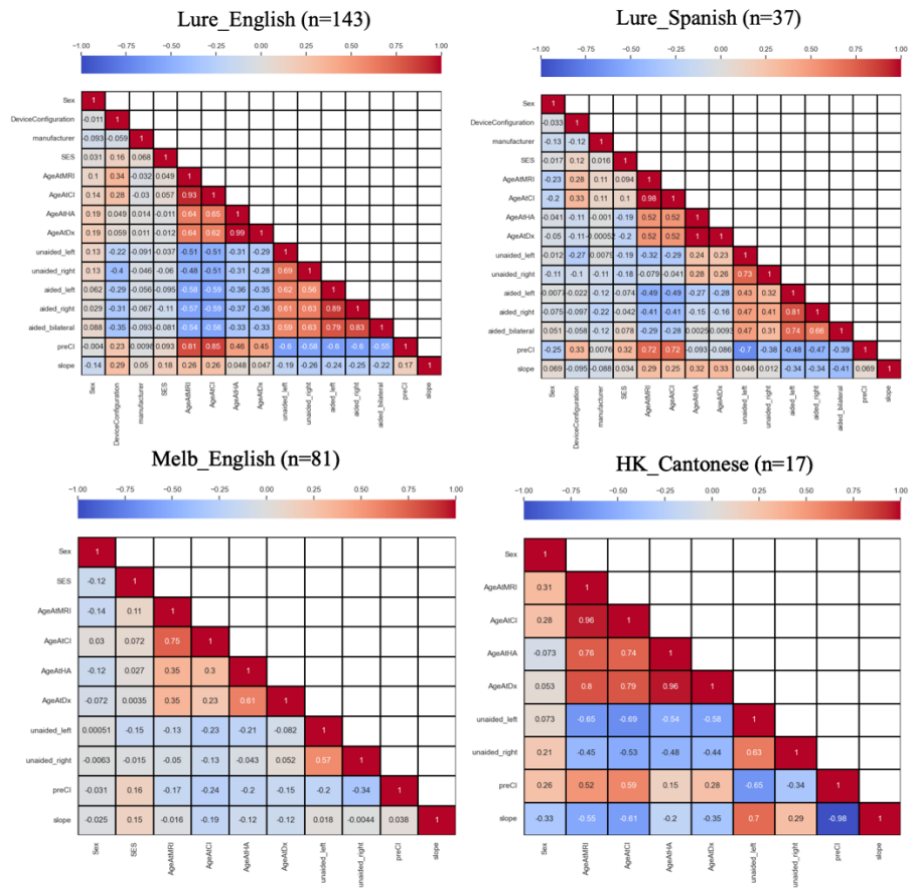

Figure S1. Correlation matrices of clinical variables across each center. Abbreviations: SES, socioeconomic status; unaided_left/right, unaided pure tone average residual hearing in the left/right ear; preCI, baseline spoken language scores; slope, fitted slope of language scores across time points.

**MRI acquisition**

*Chicago MRI data acquisition.* The T1-weighted images were obtained on a 3T Siemens scanner (MAGNETOM Skyra, Vida) using a magnetization-prepared rapid gradient-echo (MPRAGE) sequence or on a 3T General Electric MR scanner (DISCOVERY MR750, SIGNA Architect) using a 3D brain volume (BRAVO) sequence or a FSPGR (Fast Spoiled Gradient Recalled Echo) sequence. The scanning parameters were optimized to obtain a good signal-to-noise ratio (BRAVO, N=43: TE = 2.72 ms~3.91 ms, TR = 7.40 ms~9.45 ms, flip angle = 12°, matrix = 512×512, number of slices = 148~512, slice thickness = 1 mm~2 mm, voxel size = 0.3 mm×0.3 mm×0.6 mm ~ 0.5 mm×0.5 mm×1.4 mm; FSPGR, N=1: TE = 4.3 ms, TR = 10.63 ms, flip angle = 20°, matrix = 256×256, number of slices = 101, slice thickness = 1.4 mm, voxel size = 0.9 mm×0.9 mm×1.4 mm; MPRAGE, N=136: TE = 2.38 ms~3.54 ms, TR = 1490 ms~2200 ms, flip angle = 8~9°, matrix = 192×192 ~ 512×512, number of slices = 108~224, slice thickness = 0.8 mm~1 mm, voxel size = 0.8 mm×0.8 mm×0.8 mm ~ 1 mm×1 mm×1 mm).

*Melbourne MRI data acquisition.* The T1-weighted images were obtained on a 1.5T Siemens scanner (MAGNETOM Area, Avanto, and SymphonyTim) and 3T Siemens scanner (MAGNETOM Trio and Verio) using a magnetization-prepared rapid gradient-echo (MPRAGE) sequence. The scanning parameters were optimized to obtain a good signal-to-noise ratio (MPRAGE, N=81: TE = 2.31 ms~4.92 ms, TR = 11 ms~2100 ms, flip angle = 9°~20°, matrix = 576×426~224×198, number of slices = 142~452, slice thickness = 0.38 mm ~ 0.9 mm, voxel size = 0.4 mm×0.4 mm×0.8 mm ~ 0.9 mm×0.9 mm×0.9 mm).

*Hong Kong MRI data acquisition.* The T1-weighted images were acquired on a 3T Siemens Prisma scanner using a magnetization-prepared rapid gradient-echo (MPRAGE) sequence, or on a 3T General Electric MR scanner using 3D brain volume (BRAVO) sequence, or on a 3T Philips Achieva scanner using a turbo field echo (TFE) sequence. The scanning parameters were optimized to obtain a good signal-to-noise ratio (Siemens Prisma scanner: TE = 2.35 ms~2.59 ms, TR = 1800 ms, flip angle = 8°, matrix = 256×208 ~ 640×640, 192~320 slices of 0.69 mm ~ 3 mm thickness; General Electric MR scanner: TE = 2.68 ms~2.81 ms, TR = 7.62 ms~7.71 ms, flip angle = 12°, matrix = 512×512, 146~352 slices of 1 mm~1.1 mm thickness; Philips Achieva scanner: TE = 3.41 ms~3.59 ms, TR = 7.46 ms~7.77ms, flip angle = 8°, matrix = 224×224 ~ 224×280, 224~250 slices of 1.1 mm thickness).

**Voxel-based machine learning models**

We compared the classification performance of slice-based deep learning with the following voxel-based machine learning classification algorithms on the English-learning Chicago dataset: (1) Linear regression trained with L1 regularization (Lasso), L2 regularization (Ridge), and L1L2 regularization (Elastic Net). (2) Support vector machine classifier (SVM). (3) Decision tree (DT). (4) Random forest classifier (RF). (5) K-nearest neighbor classifier (KNN). (6) Gradient tree boosting-based classifier implemented in XGBoost.

Each of these popular machine learning models was trained using the direct concatenation of the preprocessed MRI data as the input.[10,11] All the methods were trained with the same preprocessed data. The dataset was divided into 80% training and validation and 20% testing. To reduce the computational cost and enhance machine learning model performance, we applied four common dimensionality reduction methods to reduce the central brain slices voxels to low-dimensional representations including principal components analysis (PCA), Gaussian Random Projection (GRP), Recursive feature Elimination (RFE), and Univariate Feature Selection (UFS). Specifically, PCA identifies orthogonal axes of maximum variance in high-dimensional data through eigen decomposition of the covariance matrix[12,13]. By projecting data onto these principal components, it achieves optimal linear dimensionality reduction while preserving global data structure. GRP employs a random matrix with entries drawn from a Gaussian distribution to project high-dimensional data into a lower-dimensional subspace. This computationally efficient method preserves pairwise distances between data points (Johnson-Lindenstrauss lemma) while introducing controlled distortion[14]. RFE iteratively trains a model, ranks features by importance, and eliminates the least significant features until a predefined feature count is reached. This wrapper method progressively refines feature subsets while maintaining predictive power[15]. Finally, UFS selects features through individual statistical tests (e.g., mutual information), ranking each feature independently against the target variable. It retains only the highest-scoring features, offering model-agnostic efficiency at the cost of ignoring feature interactions.

Consequently, a random grid search and nested cross-validation strategy were employed to validate the machine learning models and find the optimal combination of parameters for each model. Specifically, each fold was used in turn as the test set while the four remaining folds were used as training set. A grid search was utilized with five-fold cross-validation and different parameter combinations.[17] All the models were evaluated using average accuracy and average error metrics. The results were compared to determine the best model and its optimal parameters.

**Performance Evaluation Metrics**

The model's performance in classification could be evaluated using the following performance metrics: the area under the receiver operating characteristic curve (AUC), accuracy (ACC), sensitivity, and specificity. AUC measures the model's ability to discriminate between classes across various thresholds and is calculated from the False Positive Rate (FPR) and True Positive Rate (TPR). ACC measures the proportion of correctly classified images, reflecting the overall effectiveness of the model. Sensitivity, or recall, assesses the classifier's ability to correctly identify cases with the disease. Specificity evaluates how well the classifier can identify cases without the disease.

$$ACC = (TP + TN) / (TP + TN + FP + FN)$$

$$Sensitivity = TP / (TP + FN)$$

$$Specificity = TN / (FP + TN)$$

$$AUC = \int_{x=0}^{1} TPR(FPR^{-1}(x))d_x = P(X_1 > X_0)$$

where TP is true positive values, TN is true negative values, FP is false positive values, and FN is false negative values; $X_1$ is a positive instance and $X_0$ is a negative instance.

**Sensitivity Analysis**

To evaluate potential bias introduced by our slice selection strategy (i.e., slice counts or positions), we further performed extensive sensitivity analyses using alternative slice configurations. We compared model performance using three alternative 15-slice windows: superior windows from 35-50 slices encompassing the superior frontal and parietal lobes, central windows from 80-95 slices centered around the ventricular system, spanning temporal lobes, basal ganglia, and mid-ventricular region—areas consistently implicated in our targeted language-related regions, and inferior windows from 150-165 covering the brainstem, cerebellum, and inferior temporal lobes. All models were trained and evaluated using single neural network architecture, MobileNet. Results showed that the central window achieved the highest performance (AUC=0.92), outperforming both superior (AUC=0.52) and inferior (AUC=0.77). Moreover, we further evaluated the impact of slice counts using different numbers of central slices including 5, 15, and 25 slices. Results showed a slight decrease expanding to 25 slices with AUC of 0.89 and a notable drop (AUC=0.86, >6%) when reducing to 5 slices. This suggests that 15 central slices represent a reasonable trade-off between the coverage of relevant neuroanatomical structures and computational efficiency.

# References

1. Eisenberg LS, Johnson KC, Martinez AS, et al. Speech recognition at 1-year follow-up in the childhood development after cochlear implantation study: methods and preliminary findings. *Audiology and Neurotology*. 2006;11(4):259-268.
2. Wang NY, Eisenberg LS, Johnson KC, et al. Tracking development of speech recognition: longitudinal data from hierarchical assessments in the Childhood Development after Cochlear Implantation Study. *Otology & neurotology: official publication of the American Otological Society, American Neurotology Society [and] European Academy of Otology and Neurotology*. 2008;29(2):240.
3. Zimmerman IL, Steiner VG, Pond RE. Preschool Language Scale, Fifth Edition. Published online November 12, 2012. doi:10.1037/t15141-000
4. Dunn LM, Dunn DM. Peabody picture vocabulary test (4th ed.). *Circle Pines: American Guidance Service*. Published online 2007.
5. Tsiakpini L, Weichbold V, Kuehn-Inacker H, Coninx F, D'Haese P, Almadin S. *LittlEARS Auditory Questionnaire*. Austria: MED-EL; 2004.
6. Bagatto MP, Brown CL, Moodie ST, Scollie SD. External validation of the LittlEARS® Auditory Questionnaire with English-speaking families of Canadian children with normal hearing. *International journal of pediatric otorhinolaryngology*. 2011;75(6):815-817. doi:10.1016/j.ijporl.2011.03.014
7. Liu H, Jin X, Zhou Y, LI J, Liu L, NI X. Assessment and Monitoring of the LittlEARS? Auditory Questionnaire Used for Young Hearing Aid Users in Auditory Speech Development. *Journal of Audiology and Speech Pathology*. Published online 2015:291-294.
8. May-Mederake B, Kuehn H, Vogel A, et al. Evaluation of auditory development in infants and toddlers who received cochlear implants under the age of 24 months with the LittlEARS® Auditory Questionnaire. *International journal of pediatric otorhinolaryngology*. 2010;74(10):1149-1155.
9. Obrycka A, Lorens A, García JLP, Piotrowska A, Skarzynski H. Validation of the LittlEARS Auditory Questionnaire in cochlear implanted infants and toddlers. *International Journal of Pediatric Otorhinolaryngology*. 2017;93:107-116.
10. Wen J, Thibeau-Sutre E, Diaz-Melo M, et al. Convolutional neural networks for classification of Alzheimer's disease: Overview and reproducible evaluation. *Medical image analysis*. 2020;63:101694.
11. Zhao Z, Chuah JH, Lai KW, et al. Conventional machine learning and deep learning in Alzheimer's disease diagnosis using neuroimaging: A review. *Frontiers in computational neuroscience*. 2023;17:1038636.
12. Feng G, Ingvalson EM, Grieco-Calub TM, et al. Neural preservation underlies speech improvement from auditory deprivation in young cochlear implant recipients. *PNAS*. 2018;115(5):E1022-E1031. doi:10.1073/pnas.1717603115
13. Tuckute G, Sathe A, Srikant S, et al. Driving and suppressing the human language network using large language models. *Nature Human Behaviour*. Published online 2024:1-18.

**Table S1**. Performance comparison of four dimensionality-reduction techniques for feature extraction in Machine Learning Models

| Feature extraction | Model | % (95% CI) | | | AUC (95% CI) |
|---|---|---|---|---|---|
| | | Accuracy | Sensitivity | Specificity | |
| PCA | Lasso | 55.71 (49.56-61.87) | 58.57 (49.28-67.77) | 52.86 (42.74-62.97) | 0.557 (0.496-0.619) |
| | Ridge | 56.07 (51.75-60.39) | 42.86 (31.99-53.72) | 69.29 (57.30-81.27) | 0.561 (0.517-0.604) |
| | DT | 49.64 (42.22-57.06) | 37.14 (10.98-63.30) | 62.14 (47.57-76.72) | 0.496 (0.422-0.571) |
| | SVM | 47.85 (26.68-69.03) | 46.43 (24.93-67.93) | 49.29 (27.70-70.87) | 0.479 (0.267-0.690) |
| | KNN | 56.79 (50.63-62.94) | 48.57 (42.62-54.52) | 65.00 (55.39-74.61) | 0.568 (0.506-0.629) |
| | RF | 49.29 (46.31-52.26) | 43.57 (31.27-55.88) | 55.00 (44.79-65.21) | 0.493 (0.463-0.523) |
| | Xgboost | 53.93 (46.68-61.18) | 50.00 (31.45-68.55) | 57.86 (48.24-67.47) | 0.539 (0.467-0.612) |
| GRP | Lasso | 55.72 (50.91-60.52) | 53.57 (34.50-72.64) | 57.86 (37.44-78.28) | 0.557 (0.509-0.605) |
| | Ridge | 56.79 (51.98-61.59) | 52.14 (25.06-79.23) | 61.43 (42.51-80.35) | 0.568 (0.520-0.616) |
| | DT | 53.21 (45.01-61.42) | 47.86 (22.85-72.86) | 58.57 (27.59-89.55) | 0.532 (0.450-0.614) |
| | SVM | 54.64 (49.79-59.50) | 43.57 (20.23-66.91) | 65.71 (48.37-83.06) | 0.546 (0.498-0.595) |
| | KNN | 53.93 (43.84-64.02) | 50.71 (38.41-63.02) | 57.14 (46.28-68.01) | 0.539 (0.438-0.640) |
| | RF | 51.79 (48.28-55.29) | 43.57 (37.79-49.35) | 60.00 (51.47-68.53) | 0.518 (0.483-0.553) |
| | Xgboost | 55.00 (49.48-60.52) | 50.71 (39.67-61.76) | 59.29 (56.85-61.17) | 0.550 (0.495-0.605) |
| RFE | Lasso | 56.79 (51.98-61.59) | 57.86 (47.27-68.44) | 55.71 (53.29-58.14) | 0.568 (0.520-0.616) |
| | Ridge | 55.72 (50.66-60.77) | 56.43 (43.73-69.13) | 55.00 (51.03-58.97) | 0.557 (0.507-0.608) |
| | DT | 49.29 (37.81-60.76) | 38.57 (25.87-51.27) | 60.00 (36.67-83.34) | 0.493 (0.378-0.608) |
| | SVM | 51.07 (48.54-53.60) | 47.14 (42.29-52.00) | 55.00 (48.27-61.73) | 0.511 (0.485-0.536) |
| | KNN | 54.64 (49.54-59.75) | 49.29 (42.71-55.86) | 60.00 (52.71-67.29) | 0.546 (0.495-0.597) |
| | RF | 48.93 (37.13-60.72) | 46.64 (36.51-56.34) | 51.43 (36.86-66.00) | 0.489 (0.371-0.607) |
| | Xgboost | 51.07 (43.96-58.19) | 43.57 (25.18-61.96) | 58.57 (51.15-65.99) | 0.511 (0.440-0.585) |
| UFS | Lasso | 58.57 (53.76-63.38) | 51.43 (44.01-58.85) | 65.71 (61.75-69.68) | 0.586 (0.538-0.634) |
| | Ridge | 62.14 (59.25-65.03) | 55.72 (47.07-64.36) | 68.57 (61.28-75.59) | 0.621 (0.593-0.650) |
| | DT | 60.71 (56.57-64.86) | 42.86 (31.99-53.72) | 78.57 (64.20-92.94) | 0.607 (0.566-0.649) |
| | SVM | 60.36 (57.47-63.25) | 55.71 (53.29-58.14) | 65.00 (59.22-70.78) | 0.604 (0.575-0.632) |
| | KNN | 59.64 (54.30-64.98) | 56.43 (45.84-67.02) | 62.86 (54.80-70.91) | 0.596 (0.543-0.650) |
| | RF | 59.64 (54.30-64.98) | 47.86 (38.66-57.05) | 71.43 (68.29-74.56) | 0.596 (0.543-0.650) |
| | Xgboost | 59.64 (49.65-69.63) | 55.00 (42.22-67.78) | 64.29 (51.36-77.21) | 0.596 (0.497-0.696) |